\documentclass{article} 
\usepackage{collas2022_conference,times}

\usepackage[export]{adjustbox}

\usepackage{cancel,soul}
\usepackage{arydshln}
\usepackage{multirow}
\usepackage{subfig, subfloat}

\newcommand\Tstrut{\rule{0pt}{2.2ex}}         
\usepackage{amsmath}
\usepackage{xcolor}
\usepackage{bm}
\usepackage{microtype}
\usepackage{graphicx}
\usepackage{caption}
\usepackage{booktabs} 
\usepackage{amsthm}
\usepackage{wrapfig} 

\usepackage{algorithm}
\usepackage{algpseudocode}

\usepackage{amsmath,amsfonts,bm}

\def\eqref#1{equation~\ref{#1}}

\def\1{\bm{1}}

\def\vmu{{\bm{\mu}}}

\def\va{{\bm{a}}}
\def\vb{{\bm{b}}}

\def\ve{{\bm{e}}}

\def\vh{{\bm{h}}}

\def\vx{{\bm{x}}}
\def\vy{{\bm{y}}}
\def\vz{{\bm{z}}}

\def\mS{{\bm{S}}}

\def\mW{{\bm{W}}}
\def\mX{{\bm{X}}}

\DeclareMathAlphabet{\mathsfit}{\encodingdefault}{\sfdefault}{m}{sl}
\SetMathAlphabet{\mathsfit}{bold}{\encodingdefault}{\sfdefault}{bx}{n}

\DeclareMathOperator*{\argmax}{arg\,max}

\usepackage{hyperref}
\hypersetup{
    colorlinks=true,
    linkcolor=red,
    filecolor=magenta,      
    urlcolor=blue,
    citecolor=purple,
    pdftitle={Overleaf Example},
    pdfpagemode=FullScreen,
    }

\title{A Multi-Head Model for Continual Learning via Out-of-Distribution Replay}

\author{Gyuhak Kim, Zixuan Ke, Bing Liu \\ 
Department of Computer Science, University of Illinois at Chicago \\
\texttt{\{gkim87,zke4,liub\}@uic.edu} \\
}

\collasfinalcopy

\begin{document}

\maketitle
\begin{abstract}
    This paper studies \textit{class incremental learning} (CIL) of continual learning (CL).  Many approaches have been proposed to deal with catastrophic forgetting (CF) in CIL. Most methods incrementally construct a single classifier for all classes of all tasks in a single head network. To prevent CF, a popular approach is to memorize a small number of samples from previous tasks and replay them during  training of the new task. However, this approach still suffers from serious CF as the parameters learned for previous tasks are updated or adjusted with only the limited number of saved samples in the memory. This paper proposes an entirely different approach that builds a separate classifier (head) for each task (called a multi-head model) using a transformer network, called MORE. Instead of using the saved samples in memory to update the network for previous tasks/classes in the existing approach, MORE leverages the saved samples to build a task specific classifier (adding a new classification head) without updating the network learned for previous tasks/classes. The model for the new task in MORE is trained to learn the classes of the task and also to detect samples that are not from the same data distribution (i.e., \textit{out-of-distribution} (OOD)) of the task. This enables the classifier for the task to which the test instance belongs to produce a high score for the correct class and the classifiers of other tasks to produce low scores because the test instance is not from the data distributions of these classifiers. Experimental results show that MORE outperforms state-of-the-art baselines and is also naturally capable of performing OOD detection in the continual learning setting.\footnote{The code of MORE is available at \url{https://github.com/k-gyuhak/MORE}.}
\end{abstract}
\section{Introduction}
Continual learning (CL) is a learning paradigm in which a system learns a sequence of tasks sequentially and accumulates the knowledge learned in the process~\citep{ChenAndLiubook2018}. 
A main challenge of CL is how to adapt the existing knowledge in learning the new task without causing catastrophic forgetting (CF)~\citep{McCloskey1989}. CF refers to the phenomenon that the system forgets some of the previous knowledge after learning the new task due to modifications to model parameters learned for previous tasks. This paper proposes a novel method for the challenging CL setting of \textit{class incremental learning} (CIL)~\citep{Rebuffi2017,van2019three}. In this setting, the system learns a sequence of tasks $<1, \cdots, k, \cdots>$ incrementally with dataset $\mathcal{D}^{k}: \mathcal{X}^{k} \times \mathcal{Y}^k$ for each new task, which has a set of classes that are different from all the previous tasks. At any time, the system is expected to be able to classify a test instance $\vx$ to one of the classes that have been learned so far without any information about the task it belongs to.

The proposed approach is a memory-based method (also called a \textit{replay-based} method). In this method, the system saves a small fraction of training samples in a memory buffer of a fixed size after learning each task. 
Existing memory-based methods typically train a single head network, in which the network has only a single classifier for all tasks~\citep{Rebuffi2017}.
In training the new task, the algorithm replays the samples in the memory to prevent forgetting of the previously learned knowledge through joint training 
with
the current task data.
The approach learns the new task by adjusting the network parameters for previous tasks/classes while also regularizing the changes in parameters so that the knowledge learned for the previous tasks is not forgotten. However, with only a limited number of saved samples to help overcome CF, the resulting network still can forget a great deal.

This paper proposes a novel method called a \textit{M}ulti-head model for continual learning via \textit{O}OD \textit{RE}play (MORE). Unlike the existing CIL methods, MORE constructs a multi-head model which consists of a set of separate classifiers built for different tasks.
The parameters for one task in this setting do not interfere with parameters of other tasks as they are independent of each other, although there is a great deal of parameter sharing cross tasks. By construction, a multi-head model is robust to forgetting because MORE makes use of the highly effective \textit{hard attention} (also called masking) mechanism in HAT~\citep{serra2018overcoming} to protect important model parameters of each task to prevent modification and forgetting. However, HAT was originally proposed to work with the \textit{task-incremental learning} (TIL) setting of CL, which requires the correct task identifier (task-id) for each test sample to be given at inference. 
The proposed technique enables our
multi-head model/classifier to function without giving the task-id for each test sample.

The key novelty of MORE is that it utilizes the saved samples in the memory buffer entirely differently. Instead of replaying these samples to prevent forgetting in learning the new classes by updating the network learned for previous tasks or classes, MORE uses these saved samples to learn a new and independent classifier that is capable of detecting out-of-distribution (OOD) samples. 
MORE also uses a pre-trained transformer network for improved performance.\footnote{Using a pre-trained network is justified as such networks (or feature extractors) are increasingly used in computer vision as in natural language processing, where almost every application uses a pre-trained network. Furthermore, neuroscience research has shown that humans are born with rich feature detectors gained through millions of years of evolution~\citep{zhaoping2014understanding}.} 
To ensure that the pre-trained model does not see any similar data used in continual learning, in our experiments we manually remove all the data used in pre-training that may overlap with the continual learning data (see Section~\ref{sec.pre-train}).

Inspired by~\citep{houlsby2019parameter_adapter,ke2021achieving} in natural language processing, MORE does not modify the pre-trained network but only uses a trainable \textit{adapter module} inserted at each transformer layer. In continual learning, the system trains a new task specific classifier {in the shared adapter} to classify the classes of the task and also determine whether a sample is OOD. {The network parameters learned the previous tasks is left untouched in backpropagation as they are protected by trained hard attentions. 
The resulting classifier can produce a high score (e.g., prediction probability) for an in-distribution (IND) sample (the distribution of the current task training data) as it is one of the classes that the classifier has been trained for and a low score for an OOD sample as it does not belong to the IND classes of the task.} 
These can be achieved because the network is exposed to
both IND data (the current task data) and OOD data (the past data saved in the memory buffer) in training each task. 
The prediction for each test sample is made by comparing the output scores of all the task classifiers. Thus, the system does not need to know the task-id of a test instance. This base method is further improved by a Mahalanobis distance based technique (see below). 

In summary, this paper makes the following contributions.
\begin{itemize}
    \item It proposes a novel approach for class incremental learning (CIL). It builds a multi-head model and trains each task using samples in the replay memory as OOD data against the current task data 
    unlike the
    existing methods.
    To the best of knowledge, this way of using the memory or replay data has not been done before.
    \item It further improves the above base method by using a Mahalanobis distance-based technique.
    The paper
    proposes to combine this distance factor and the softmax probability output scores of the above base method to make the final prediction decision. This method can be considered as an ensemble of the two methods.
    \item Since the proposed method is learned based on OOD detection for each task, it can naturally perform OOD detection in the continual learning (CL) setting to identify any OOD test instances that do not belong to any of the learned tasks so far. Again, we are unaware of any existing CL system that does continual OOD detection.
\end{itemize}
Experimental results show that the proposed method outperforms existing state-of-the-art CIL method both in terms of IND classification accuracy and continual OOD detection.

\section{Related Work}
A large number of approaches have been proposed to prevent forgetting in continual learning (see the book~\citep{ChenAndLiubook2018} and the survey~\citep{masana2020class}). Using regularizations~\citep{Kirkpatrick2017overcoming} is a popular approach. This approach approximates the importance of parameters for the previous tasks, and penalizes changes to them to prevent forgetting~\citep{Kirkpatrick2017overcoming, Zenke2017continual, ritter2018online, Dhar2019CVPR, mirzadeh2021linear}. Our approach is different as we do not use regularization.

Saving a small fraction of training data in a memory and replaying it to prevent forgetting is another very popular approach, called the \textit{replay} method. The saved samples are used to distill the knowledge of previous tasks~\citep{Rusu2016, Rebuffi2017, hou2019learning, wu2019large, NEURIPS2020_b704ea2c_derpp, yan2021dynamically} or replayed to restrict the loss from increasing on the memory or selective representations~\citep{Lopez2017gradient, chaudhry2018efficient, Chaudhry_Gordo_Dokania_Torr_Lopez-Paz_2021_hal}. Some researchers also proposed different replay strategies~\citep{riemer2018learning, aljundi2019gradient, chaudhry2019continual_er, Liu_2020_CVPR}. Our method MORE also uses a memory, but it does not replay the samples to regularize the adjustments to the previous task model parameters in learning the current task so that the previous task models are minimally changed. MIR~\citep{NIPS2019_9357_mir} proposes a sampling strategy that draws maximally interfering samples from memory. MORE does not propose a sampling strategy, but uses the saved samples as OOD data to improve the current task learning. Models of the previous tasks are not updated. Some methods~\citep{lee2019overcoming, Liu2020, smith2021memory} use external data to improve knowledge distillation for overcoming forgetting. Our method does not use any external data or knowledge distillation.

\textit{Pseudo-replay} is another popular approach~\citep{Shin2017continual}, which
constructs a separate encoder-decoder network to generate raw data similar to previous task data. During training a new task, the system generates pseudo data similar to the samples of previous tasks. The generated data and new task data are trained together to distinguish the previous and current classes \citep{Kamra2017deep, Shin2017continual, ostapenko2019learning, Rostami2019ijcai, van2020brain, Zhu_2021_CVPR_pass}. Our method MORE does not use this approach.

The list of other approaches is also extensive. \textit{Parameter isolation} methods~\citep{serra2018overcoming, supsup2020, henning2021posterior} are highly effective at preventing forgetting in task incremental learning, which requires the task-id for each test instance during inference. MORE borrows the hard attention~\citep{serra2018overcoming} idea to prevent modifications to each previous task model. However, it does not need the task-id in testing and further uses memory data for OOD detection to improve CIL performances. \textit{Gradient projection}~\citep{zeng2019continuous} projects gradients to a subspace orthogonal to previously learned parameter space to prevent intervention. Some such methods~\citep{kim2020continual, chaudhry2020continual, saha2021gradient} require task-id for more complex tasks to recall the task-specific projection at inference.
PCL~\citep{hu2021continual_pcl} constructs an individual classifier for each class by using a fixed pre-trained feature extractor. We build an OOD model for each task and use hard attention to protect it.

A related line of research is online CL. Contrary to batch CL where the system sees the full training data when the task arrives and can train any number of epochs, online CL only sees the training data once (training in one epoch).
There are many existing online CL systems, e.g., SLDA~\citep{Hayes_2020_CVPR_Workshops},
REMIND~\citep{hayes2020remind} and OCM~\citep{guo2022online}.
\citep{Roady_2020_Stream51} also proposed a new dataset. Our method is a batch CL method.

Recent approaches in task incremental learning (where the task id is required for each test instance) have attempted to use a multi-head model for CIL.
CCG~\citep{abati2020conditional} constructs an additional network to predict the task-id.
MORE does not build another network.
iTAML~\citep{rajasegaran2020itaml} uses the correct task network by identifying the task-id of the test data in a batch. The limitation is that it requires the test data in a batch to belong to the same task.
MORE is different as it can predict for a single instance at a time. HyperNet~\citep{von2019continual_hypernet} and PMCL~\citep{henning2021posterior} propose an entropy based task-id prediction method.
Our method enables each task model to produce better scores on OOD outputs and should also be applicable to HyperNet and PMCL. SupSup~\citep{supsup2020} predicts task-id
by finding the optimal superpositions at inference. Our method does not require any additional operations such as entropy or optimization to predict task-id. It simply chooses the class with the maximum probability among all task model outputs. Our recent work CLOM~\citep{Kim_2022_CVPR} uses contrastive learning and data augmentation to build an OOD classifier for each task, but it is not a replay method and is weaker than MORE.

\section{Proposed MORE Technique}
As mentioned earlier, existing class incremental learning (CIL) methods mostly train
a single-head network for all the classes learned so far (i.e. a single classifier for all the tasks)~\citep{van2019three}.
To learn each task, the parameters for previous tasks must be modified, which causes CF. To alleviate CF, the popular replay approach saves a small number of samples of previous tasks in a memory/replay buffer and use them in learning the new task.
However, this results in a biased network due to sample imbalance between the saved samples and current data~\citep{Rebuffi2017}. We take an entirely different approach by training a multi-head network as an adapter to a pre-trained network. Each head is an OOD detection model for a task. Hard attentions are employed to protect each task model or classifier.

\subsection{Training an OOD Detection Model}\label{sec:training_ood}
\begin{figure}
\centering
\includegraphics[width=.83\linewidth]{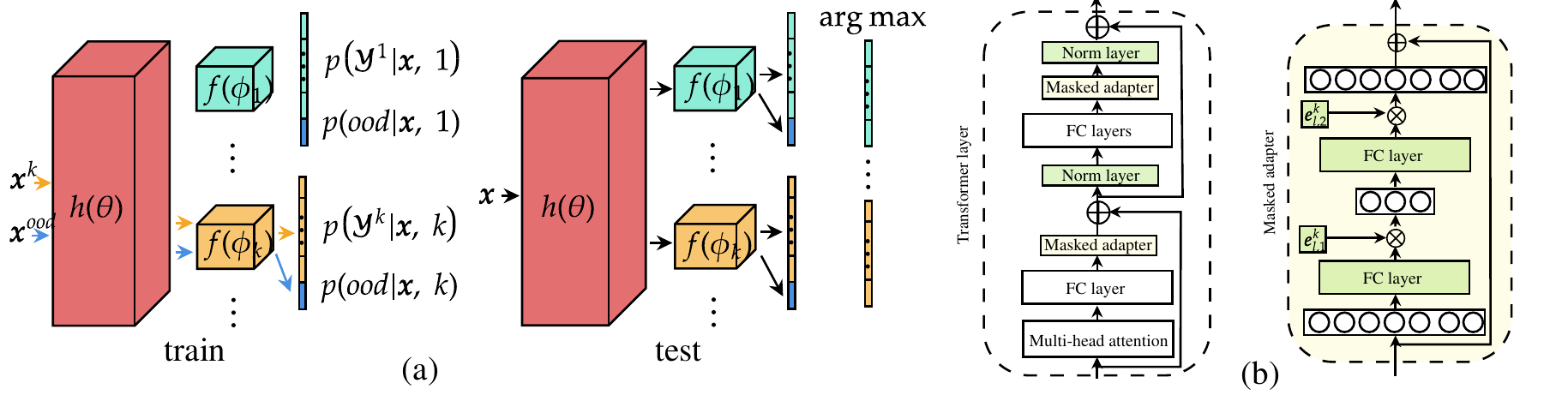}
\caption{
(a) We train the feature extractor and the task classifier $k$ at task $k$.
The output values of the classifier correspond to $|\mathcal{Y}^{k}| + 1$ classes, in which the last class is for OOD (i.e., representing previous and unseen future classes).
At inference/testing, the probability values of each task model without the OOD class are concatenated and the system chooses the class with the maximum score. (b) Transformer and adapter module. The masked adapter network consists of 2 fully connected layers and task specific masks. During training, only the masked adapters and norm layers are updated and the other parts in the transformer layers remain unchanged.
}
\label{illustration}
\end{figure}

At task $k$, the system receives the training data $\mathcal{D}^{k}: \mathcal{X}^{k} \times \mathcal{Y}^k$. We train the feature extractor $\vz = h(\vx, k; \theta)$ and task specific classifier $f(\vz; \phi_k)$ using $\mathcal{D}^{k}$ and the samples in the memory buffer $\mathcal{M}$. We treat the buffer data as OOD data to encourage the network to learn the current task and also detect ODD samples (the models or classifiers of the previous tasks are not touched). We achieve it by maximizing $p(y | \vx, k) = \text{softmax} f(h(\vx, k; \theta); \phi_k)$ for an IND sample $\vx \in \mathcal{X}^{k}$ and maximizing $p(ood|\vx, k)$ for an OOD sample $\vx \in \mathcal{M}$. The additional label $ood$ is reserved for previous and future unseen classes.
Figure~\ref{illustration}(a) shows the overall idea of the proposed approach. We formulate the problem as follows.

Given the training data $\mathcal{D}^{k}$
of size $N$ at task $k$ and the memory buffer $\mathcal{M}$ of size $M$, we minimize the loss
\begin{align}
    \mathcal{L}_{ood} (\theta, \phi_k) = - \frac{1}{M + N} \sum_{(\vx, y) \in \mathcal{M}} \log p(ood | \vx, k) + \sum_{(\vx, y) \in \mathcal{D}^{k}} \log p(y | \vx, k) \label{ood_obj}
\end{align}
It is the sum of two categorical cross-entropy losses. The first loss is for learning OOD samples while the second loss is for learning the classes from the current task. We optimize the shared parameter $\theta$ in the feature extractor. The task specific classification parameters $\phi_k$ are independent of other tasks. 
The learned representation on the current data is robust to outliers or OOD data. The classifier produces more accurate outputs on both IND and OOD data.

In testing,
we perform prediction by comparing the softmax probability output values using all the task classifiers from task 1 to k without the OOD class as
\begin{align}
    \hat{y} = \argmax \bigoplus_{1 \leq j \leq k} p(\mathcal{Y}^{j}| \vx, j) \label{base_prediction}
\end{align}
where $\bigoplus$ is the concatenation over the output space. Figure~\ref{illustration}(a) shows the prediction rule.
We are basically choosing the class with the highest softmax probability over each task among all the learned classes.

\subsection{Back-Updating the Previous OOD Models}\label{sec:backward}
Each task model works better if more diverse OOD data is provided during training. As in a replay-based approach, MORE saves an equal number of samples per class after each task~\citep{chaudhry2019continual_er}. The saved samples in the memory are used as OOD samples for each new task. 
Thus, in the beginning of continual learning when the system is trained on only a small number of tasks, the classes of samples in the memory are less diverse than after more tasks are learned. This makes the performance of OOD detection stronger for later tasks, but weaker in earlier tasks. To prevent this asymmetry, we update the model of each previous task so that it can also identify the samples from subsequent classes (which were unseen during the training of the previous task) as OOD samples.

At task $k$, we update the previous task models $(j=1, \cdots, k-1)$ as follows. Denote the samples of task $j$ in memory $\mathcal{M}$ by $\tilde{\mathcal{D}}^{j}$. We construct a new dataset using the current task dataset and the samples in the memory buffer. We randomly select $|\mathcal{M}|$ samples from the training data $\mathcal{D}^{k}$ and combine it with the remaining samples in $\mathcal{M}$ after removing the IND samples $\tilde{\mathcal{D}^{j}}$ of task $j$. We do not use the entire training data $\mathcal{D}^{k}$ as we do not want a large sample imbalance between IND and OOD.
Denote the new dataset by $\tilde{\mathcal{M}}$.
Using the data, we update only the parameters $\phi_j$ of the classifier for task $j$ with the feature representations frozen by minimizing the loss
\begin{align}
    \mathcal{L}(\phi_{j}) = - \frac{1}{2M} \sum_{(\vx, y) \in \tilde{\mathcal{M}}} \log p(ood | \vx, j) + \sum_{(\vx, y) \in \tilde{\mathcal{D}}^{j}} \log p(y | \vx, j) \label{backward}
\end{align}
We reduce the loss by updating the parameters of classifier $j$ to maximize the probability of the class if the sample belongs to task $j$ and maximize the OOD probability otherwise.

\subsection{Improving Prediction Performance by a Distance Based Technique}\label{sec:ensemble_scores}
We further improve the prediction in Eq.~\ref{base_prediction} by introducing a distance based factor used as a coefficient to the softmax probabilities in Eq.~\ref{base_prediction}. It is quite intuitive that if a test instance is close to a class, it is more likely to belong to the class. We thus propose to combine this new distance factor and the softmax probability output of the task $k$ model to make the final prediction decision. In some sense, this can be considered as an ensemble of the two methods.  

We define the distance based coefficient $s^{k}(\vx)$ of task $k$ for the test instance $\vx$ by the maximum of inverse Mahalanobis distance~\citep{lee2018simple} between the feature of $\vx$ and the Gaussian distributions of the classes in task $k$ parameterized by the mean $\vmu^k_j$ of the class $j$ in task $k$ and the sample covariance $\mS^{k}$. They are estimated by the features of class $j$'s training data for each class $j$ in task $k$.
If a test instance is from the task, its feature should be close to the distribution that the instance belongs to. Conversely, if the instance is OOD to the task, its feature should not be close to any of the distributions of the classes in the task.
More precisely, for task $k$ with class $y_1, \cdots, y_{|\mathcal{Y}^{k}|}$ (where ${|\mathcal{Y}^{k}|}$ represents the number of classes in task $k$), we define the coefficient $s^{k}(\vx)$ as
\begin{align}
    s^{k}(\vx) = \max \left[ c/\text{MD}(\vx; \vmu^k_{y_{1}}, \mS^{k}), \cdots, c/\text{MD}(\vx; \vmu^k_{y_{1 + |\mathcal{Y}^{k}|}}, \mS^{k}) \right] \label{eq:ensemble}
\end{align}
{where $c$ is a positive constant and $\text{MD}(\vx; \vmu^k_{j}, \mS^{k})$ is the Mahalanobis distance.
The coefficient is large if at least one of the Mahalanobis distances is small but the coefficient is small if all the distances are large (i.e. the feature is far from all the distributions of the task).
The parameters $\vmu^k_{j}$ and $\mS^{k}$ can be computed and saved when each task is learned. The mean $\vmu^k_{j}$
is computed using the training samples $\mathcal{D}_j^{k}$ of class $j$ as follows,
\begin{align}
    \vmu^k_{j} = \sum_{\vx \in \mathcal{D}_{j}^{k}} h(\vx, j) / |\mathcal{D}_{j}^{k}| \label{eq:compute_mu}
\end{align}
and the covariance $\mS^{k}$ of task $k$ is the mean of covariances of the classes in task $k$,
\begin{align}
    \mS^{k} = \sum_{j \in \mathcal{Y}^{k}} \mS_{j}^{k} / |\mathcal{Y}^{k}| \label{eq:compute_sigma}
\end{align}
where $\mS_{j}^{k} = \sum_{\vx \in \mathcal{D}_{j}^{k}} (\vx - \vmu^k_{j})^{T}(\vx - \vmu^k_{j}) / |\mathcal{D}_{j}^{k}|$ is the sample covariance of class $j$. 
By multiplying the coefficient $s^{k}(\vx)$ to the original softmax probabilities $p(\mathcal{Y}^{k} | \vx, k)$, the task output $p(\mathcal{Y}^{k} | \vx, k) s^{k}(\vx)$ increases if $\vx$ is from task $k$ and decreases otherwise. The final prediction is made by (which replaces Eq.~\ref{base_prediction})
\begin{align}
    y = \argmax \bigoplus_{1\leq k \leq t}  p(\mathcal{Y}^{k} | \vx, k) s^{k}(\vx), \label{final_prediction}
\end{align}
}

\subsection{Preventing Forgetting by Hard Attention Masks}
We have discussed how to train a task model
capable of CIL prediction as an OOD classifier. In this section, we use notations from \citep{Kim_2022_CVPR} and explain how to use the hard attention mechanism~\citep{serra2018overcoming} to protect the feature extractor of each task model. In learning a task, a set of embeddings are trained to protect the important neurons so that the corresponding parameters are not interfered by subsequent tasks. The importance of a neuron is measured by the 0-1 pseudo-step function, where 0 indicates important and 1 indicates not important (and thus trainable).

The hard attention mask is an output of sigmoid function $u$ with a hyper-parameter $s$ as
\begin{align}
    \va_{l}^{k} = u(s \ve_{l}^{k}), \label{eq:smax}
\end{align}
where
$\ve_{l}^{k}$ is a learnable embedding at layer $l$ of task $k$. Since the step function is not differentiable, a sigmoid function with a large $s$ is used to approximate it. Sigmoid is approximately a 0-1 step function with a large $s$. The attention is multiplied to the output $\vh_{l} = \text{ReLU}(\mW_l \vh_{l-1} + \vb_l)$ of layer $l$,
\begin{align}
    \vh'_{l} = \va_{l}^{k} \otimes \vh_{l}
\end{align}
The $j$th element $a_{j, l}^{k}$ in the attention mask blocks (or unblocks) the information flow from neuron $j$ at layer $l$ if its value is $0$ (or $1$). With 0 value of $a_{j, l}^{k}$, the corresponding parameters in $\mW_{l}$ and $\vb_{l}$
can be freely changed as the output values $\vh'_{l}$ are not affected. The neurons with non-zero mask values are necessary to perform the task, and thus need a protection for catastrophic forgetting.

We modify the gradients of parameters that are important in performing the previous tasks $(1, \cdots, k-1)$ during training task $k$ so they are not interfered. Denote the accumulated mask by
\begin{align}
    \va_{l}^{<k} = \max(\va_{l}^{<k-1}, \va_{l}^{k-1})
\end{align}
where $\max$ is an element-wise maximum and the initial mask $\va_{l}^{0}$ is a zero vector. It is a collection of mask values at layer $l$ where a neuron has value 1 if it has ever been activated previously.
The gradient of parameter $w_{ij, l}$ is modified as
\begin{align}
    \nabla w_{ij, l}' = \left( 1 - \min\left( a_{i,l}^{<k}, a_{j, l-1}^{<k} \right) \right) \nabla w_{ij, l} \label{eq:grad_mod}
\end{align}
where $a_{i,l}^{<k}$ is the $i$th unit of $\va_{l}^{<t}$. The gradient flow is blocked if both neurons $i$ in the current layer and $j$ in the previous layer have been activated. We apply the mask for all layers except the last layer. The parameters in last layer do not need to be protected as they are task-specific parameters.

A regularization is introduced to encourage sparsity in $\va_{l}^{k}$ and parameter sharing with $\va_{l}^{<k}$. The capacity of a network depletes when $\va_{l}^{<k}$ becomes 1-vector in all layers. Despite a set of new neurons can be added in network at any point in training for more capacity, we utilize resources more efficiently by minimizing the loss
\begin{align}
    \mathcal{L}_r = \lambda \frac{\sum_l \sum_i a_{i,l}^{k}\left( 1 - a_{i,l}^{<k} \right)}{\sum_{l} \sum_{i} \left(1 - a_{i, l}^{<k} \right)} \label{eq:hat_reg}
\end{align}
where $\lambda$ is a hyper-parameter. The final objective to train a comprehensive task network without forgetting is
\begin{align}
    \mathcal{L} = \mathcal{L}_{ood} + \mathcal{L}_{r} \label{final_obj}
\end{align}
where $\mathcal{L}_{ood}$ is Eq.~\ref{ood_obj} for training an OOD model discussed in Sec.~\ref{sec:training_ood} after incorporating the trainable embedding $\ve^{k}$. We describe the whole training and prediction process in Appendix~\ref{appendix:pseudo}.

\section{Experiments}
\textbf{Evaluation Datasets}: We use three image classification benchmark datasets in our experiments.\footnote{We conduct additional experiment using SVHN (very different from pre-training data) and include the result in Appendix~\ref{appendix:additional_data}.}

(1). \textbf{CIFAR-10}~\citep{Krizhevsky2009learning}
: This is an image classification dataset consisting of 60,000 32x32 color images of 10 classes, with 6,000 images per class. We use 5,000 samples for training and 1,000 samples for testing.

(2). \textbf{CIFAR-100}~\citep{Krizhevsky2009learning}
: This dataset consists of 60,000 32x32 color images of 100 classes, with 600 images per class. We use 500 samples for training and 100 for testing.

(3). \textbf{Tiny-ImageNet}~\citep{Le2015TinyIV}:
This dataset consists of 120,000 64x64 color images of 200 classes, with 600 images per class, in which 500, 50, and 50 images are for training, validation, and testing, respectively.
Following the previous works~\citep{Zhu_2021_CVPR_pass}, we use the validation data for testing as the test data has no labels.

\textbf{Baseline Systems}: We compare our method MORE with 6 state-of-the-art baselines. For gradient modification methods, we consider \textbf{OWM}~\citep{zeng2019continuous}. For replay-based methods, we compare against \textbf{iCaRL}~\citep{Rebuffi2017}, \textbf{A-GEM}~\citep{chaudhry2018efficient}, \textbf{EEIL}~\citep{castro2018end}, \textbf{GD}~\citep{lee2019overcoming} without external data, \textbf{DER++}~\citep{NEURIPS2020_b704ea2c_derpp}, and \textbf{HAL}~\citep{Chaudhry_Gordo_Dokania_Torr_Lopez-Paz_2021_hal}.
For pseudo-replay method, we include \textbf{PASS}~\citep{Zhu_2021_CVPR_pass}. For multi-head methods, we compare  with HAT~\citep{serra2018overcoming} using the task-id prediction method proposed in HyperNet~\citep{von2019continual_hypernet}.\footnote{CCG~\citep{abati2020conditional}, iTAML~\citep{rajasegaran2020itaml}, HyperNet~\citep{von2019continual_hypernet}, and SupSup~\citep{supsup2020} are not included because: CCG's code is not released.
iTAML requires test samples in a batch of the same task.
When a single test instance is provided, its accuracy on CIFAR100-10T is 33.5\%, which is much lower than many other baselines. HyperNet and SupSup cannot work with the transformer adapter due to their architectures. The CIL performances of HyperNet and SupSup with ResNet-18 on CIFAR100-10T are 29.6 and 33.1, respectively, which are even lower than that of iTAML. Our earlier system CLOM~\citep{Kim_2022_CVPR} (which is also based on OOD detection as well) is not included as it is much weaker (up to 15\% lower than MORE in accuracy) and its approach of using contrastive learning and data augmentations does not work well with a pre-trained model.} 

\subsection{Pre-trained Network}
\label{sec.pre-train}

We pre-train a vision transformer~\citep{touvron2021training_deit} using a subset of the ImageNet data~\citep{russakovsky2015imagenet} and apply the pre-trained network to all baselines and our method. To ensure that there is no overlapping of data between ImageNet and our experimental datasets, we manually removed 389 classes from the original 1000 classes in ImageNet that are similar/identical to the classes in CIFAR-10, CIFAR-100, or Tiny-ImageNet. We pre-train the network with the remaining subset of 611 classes of ImageNet.

Using the pre-trained network, both our system and the baselines improve dramatically compared to their versions without using the pre-trained network. For instance, the two best baselines (DER++ and PASS) in our experiments achieve the average classification accuracy of 66.89 and 68.25 (after the final task) with the pre-trained network over 5 experiments while they achieve only 46.88 and 32.42 without using the pre-train network. 

We insert an adapter module at each transformer layer to exploit the pre-trained transformer network in continual learning. During training, the adapter module and the layer norm are trained while the transformer parameters are unchanged to prevent forgetting in the pre-trained network. Existing continual learning methods in natural language processing commonly use pre-trained transformers and adapter-like modules~\citep{ke2021achieving}. Therefore, we also leverage it for computer vision tasks.

\subsection{Training Details}

For all experiments, {we use the same backbone architecture DeiT-S/16~\citep{touvron2021training_deit} with 2-layers adapter~\citep{houlsby2019parameter_adapter} at each transformer layer}, and the same class order for both baselines and our method. The first fully-connected layer in adapter maps from dimension 384 to bottleneck.
The second fully-connected layer following ReLU activation function maps from bottleneck to 384. The the bottleneck dimension is the same for all adapters in a model. For our method, we use SGD with momentum value 0.9.  The back-update method in Sec.~\ref{sec:backward} is also a hyper-parameter choice. If we apply it, we train each classifier for 10 epochs by SGD with learning rate 0.01, batch size 16, and momentum value 0.9. We choose 500 for $s$ in Eq.~\ref{eq:smax}, 0.75 for $\lambda$ in Eq.~\ref{eq:hat_reg} as recommended in \cite{serra2018overcoming}, and 20 for $c$ in distance based coefficient in Eq.~\ref{eq:ensemble} of Sec.~\ref{sec:ensemble_scores}.
We find a good set of learning rate and number of epochs on the validation set made of 10\% of training data.
We follow~\citep{chaudhry2019continual_er} and save an equal number of random samples per class in the replay memory. Following the experiment settings in~\citep{Rebuffi2017, Zhu_2021_CVPR_pass}, we fix the size of memory buffer and reduce the saved samples to accommodate a new set of samples after a new task is learned. We use the class order protocol in \citep{Rebuffi2017,NEURIPS2020_b704ea2c_derpp} by generating random class orders for the experiments. The baselines and our method use the same class ordering for fairness. We also report the size of memory required for each experiment in Appendix~\ref{appendix:size_of_memory}.

For CIFAR-10, we split 10 classes into 5 tasks (2 classes per task).
The bottleneck size in each adapter is 64. Following \citep{NEURIPS2020_b704ea2c_derpp}, we use the memory size 200, and train for 20 epochs with learning rate 0.005, and apply the back-update method in Sec.~\ref{sec:backward}.

For CIFAR-100, we conduct 10 tasks and 20 tasks experiments, where each task has 10 classes and 5 classes, respectively.
We double the bottleneck size of the adapter to learn more classes. We use the memory size 2000 {following \citep{Rebuffi2017}} and train for 40 epochs with learning rate 0.001 and 0.005 for 10 tasks and 20 tasks, respectively, and apply the back-update method in Sec.~\ref{sec:backward}.

For Tiny-ImageNet, two experiments are conducted. We split 200 classes into 5 and 10 tasks, where each task has 40 classes and 20 classes per task, respectively. We use the bottleneck size 128, and save 2000 samples in memory. We train with learning rate 0.005 for 15 and 10 epochs for 5 tasks and 10 tasks, respectively.
There is no need to use the back-update method as the earlier tasks already have diverse OOD classes.

\subsection{Results and Comparisons}\label{sec:result_comparison}
We evaluate our method with the standard metrics: \textit{average classification accuracy} (ACA), \textit{average incremental accuracy} (AIA), and \textit{average performance reduction rate} (which is commonly called the \textit{average forgetting rate}~\citep{Liu_2020_CVPR}) after the final task. The average classification accuracy $\mathcal{A}^k$ after task $k$ is the accuracy of all seen classes after task $k$. We report ACA after the final task. The average incremental accuracy~\citep{Rebuffi2017} $\mathcal{A}$ after the final task $t$ is defined as $\mathcal{A} = \sum_{k=1}^{t} \mathcal{A}^{k} / t$.
The average performance reduction rate is $\mathcal{R}^{t} = \sum_{k=1}^{t-1} (\mathcal{A}^{\text{init}}_{k} - \mathcal{A}^{t}_{k}) / (t-1)$, where $\mathcal{A}_{k}^{\text{init}}$ is the classification accuracy on samples of task $k$ right after learning the task $k$. We do not consider the task $t$ as it is the last task. Note that we do not use the term \textit{average forgetting rate} as in other papers~\citep{Liu_2020_CVPR} because there are actually two factors that cause the performance degradation in continual learning: \textit{forgetting} and \textit{more classes} (as the system learns more tasks/classes, the classification accuracy will naturally decrease). That is why we use the term \textit{average performance reduction rate} instead.
All the reported results are the averages of 5 runs.

\begin{table*}[t]
\caption{
Average accuracy (ACA) and average incremental accuracy (AIA) after the final task. `-XT' means X number of tasks. Our system MORE and all baselines used the pre-trained network. The last two columns show the average ACA and AIA of each method over all datasets and experiments. We highlight the best results in each column in bold.
}
\label{Tab:maintable}
\centering
\resizebox{0.96\columnwidth}{!}{
\begin{tabular}{l c c c c c c c c c c | c c}
&&&&&&&&&&&&\\[0.0em]
\toprule
\multirow{2}{*}{Method}  & \multicolumn{2}{c}{CIFAR10-5T}  &  \multicolumn{2}{c}{CIFAR100-10T} &  \multicolumn{2}{c}{CIFAR100-20T} &  \multicolumn{2}{c}{T-ImageNet-5T} & \multicolumn{2}{c}{T-ImageNet-10T} & \multicolumn{2}{|c}{Average}\\
{} & ACA & AIA & ACA & AIA & ACA & AIA & ACA & AIA & ACA & AIA & ACA & AIA \\
\midrule
OWM             &  41.69\scalebox{0.8}{$\pm$6.34}  & 59.07\scalebox{0.8}{$\pm$3.31} & 21.39\scalebox{0.8}{$\pm$3.18} & 39.71\scalebox{0.8}{$\pm$1.35} & 16.98\scalebox{0.8}{$\pm$4.44} & 32.18\scalebox{0.8}{$\pm$1.51} & 24.55\scalebox{0.8}{$\pm$2.48} & 45.65\scalebox{0.8}{$\pm$1.15} & 17.52\scalebox{0.8}{$\pm$3.45} & 35.57\scalebox{0.8}{$\pm$1.83} & 24.43 & 41.99 \Tstrut \\
iCaRL         & 87.55\scalebox{0.8}{$\pm$0.99}  & 92.75\scalebox{0.8}{$\pm$1.08}   &  68.90\scalebox{0.8}{$\pm$0.47}  & 77.82\scalebox{0.8}{$\pm$1.28}  & 69.15\scalebox{0.8}{$\pm$0.99}  & 77.74\scalebox{0.8}{$\pm$1.82}  & 53.13\scalebox{0.8}{$\pm$1.04}  & 63.35\scalebox{0.8}{$\pm$2.02}  & 51.88\scalebox{0.8}{$\pm$2.36}  & 64.62\scalebox{0.8}{$\pm$0.97}  & 66.12 & 75.26 \\ 
A-GEM         &  56.33\scalebox{0.8}{$\pm$7.77}  & 71.22\scalebox{0.8}{$\pm$1.42}   &  25.21\scalebox{0.8}{$\pm$4.00}  & 43.39\scalebox{0.8}{$\pm$0.88}  & 21.99\scalebox{0.8}{$\pm$4.01}  & 35.56\scalebox{0.8}{$\pm$0.95}  & 30.53\scalebox{0.8}{$\pm$3.99}  & 50.37\scalebox{0.8}{$\pm$2.15}  & 21.90\scalebox{0.8}{$\pm$5.52}  & 39.79\scalebox{0.8}{$\pm$3.28}  & 31.20 & 47.74 \\
EEIL & 82.34\scalebox{0.8}{$\pm$3.13} & 90.50\scalebox{0.8}{$\pm$0.72} & 68.08\scalebox{0.8}{$\pm$0.51} & 81.09\scalebox{0.8}{$\pm$0.37} & 63.79\scalebox{0.8}{$\pm$0.66} & 79.54\scalebox{0.8}{$\pm$0.69} & 53.34\scalebox{0.8}{$\pm$0.54} & 66.63\scalebox{0.8}{$\pm$0.40} & 50.38\scalebox{0.8}{$\pm$0.97} & 66.54\scalebox{0.8}{$\pm$0.61} & 63.59 & 76.86 \\
GD & \textbf{89.16}\scalebox{0.8}{$\pm$0.53} & 94.22\scalebox{0.8}{$\pm$0.75} & 64.36\scalebox{0.8}{$\pm$0.57} & 80.51\scalebox{0.8}{$\pm$0.57} & 60.10\scalebox{0.8}{$\pm$0.74} & 78.43\scalebox{0.8}{$\pm$0.76} & 53.01\scalebox{0.8}{$\pm$0.97} & 67.51\scalebox{0.8}{$\pm$0.38} & 42.48\scalebox{0.8}{$\pm$2.53} & 63.91\scalebox{0.8}{$\pm$0.40} & 61.82 & 76.92 \\
DER++         &  84.63\scalebox{0.8}{$\pm$2.91}  & 91.81\scalebox{0.8}{$\pm$0.65}   & 69.73\scalebox{0.8}{$\pm$0.99}  & \textbf{81.71}\scalebox{0.8}{$\pm$0.67}  & 70.03\scalebox{0.8}{$\pm$1.46}  & \textbf{82.24}\scalebox{0.8}{$\pm$0.79}  & 55.84\scalebox{0.8}{$\pm$2.21} & 68.47\scalebox{0.8}{$\pm$0.73} & 54.20\scalebox{0.8}{$\pm$3.28}  & 68.06\scalebox{0.8}{$\pm$1.04} & 66.89 & 78.46 \\
HAL           &  84.38\scalebox{0.8}{$\pm$2.70}  & 90.41\scalebox{0.8}{$\pm$1.04}  & 67.17\scalebox{0.8}{$\pm$1.50}  & 78.62\scalebox{0.8}{$\pm$0.45}  & 67.37\scalebox{0.8}{$\pm$1.45}  & 78.43\scalebox{0.8}{$\pm$0.61}  & 52.80\scalebox{0.8}{$\pm$2.37}  & 67.52\scalebox{0.8}{$\pm$0.93}  & 55.25\scalebox{0.8}{$\pm$3.60}  & 67.89\scalebox{0.8}{$\pm$2.32}  & 65.39 & 76.57 \\
PASS          &  86.21\scalebox{0.8}{$\pm$1.10}  & 91.78\scalebox{0.8}{$\pm$1.12}   & 68.90\scalebox{0.8}{$\pm$0.94}  & 78.27\scalebox{0.8}{$\pm$0.81}   & 66.77\scalebox{0.8}{$\pm$1.18}  & 77.01\scalebox{0.8}{$\pm$1.13}  & 61.03\scalebox{0.8}{$\pm$0.38} & 70.02\scalebox{0.8}{$\pm$0.56} & 58.34\scalebox{0.8}{$\pm$0.42}  & 68.45\scalebox{0.8}{$\pm$1.20} & 68.25 & 77.11 \\
HAT          &  83.30\scalebox{0.8}{$\pm$1.54}  & 91.06\scalebox{0.8}{$\pm$0.36}   & 62.34\scalebox{0.8}{$\pm$0.93}  & 73.99\scalebox{0.8}{$\pm$0.86}   & 56.72\scalebox{0.8}{$\pm$0.44}  & 69.12\scalebox{0.8}{$\pm$1.06}  & 57.91\scalebox{0.8}{$\pm$0.72} & 69.38\scalebox{0.8}{$\pm$1.14} & 53.12\scalebox{0.8}{$\pm$0.94}  & 65.63\scalebox{0.8}{$\pm$1.64} & 62.68 & 73.84 \\
\hline
MORE & \textbf{89.16}\scalebox{0.8}{$\pm$0.96} & \textbf{94.23}\scalebox{0.8}{$\pm$0.82} & \textbf{70.23}\scalebox{0.8}{$\pm$2.27} & 81.24\scalebox{0.8}{$\pm$1.24} & \textbf{70.53}\scalebox{0.8}{$\pm$1.09} & 81.59\scalebox{0.8}{$\pm$0.98} & \textbf{64.97}\scalebox{0.8}{$\pm$1.28}  & \textbf{74.03}\scalebox{0.8}{$\pm$1.61} & \textbf{63.06}\scalebox{0.8}{$\pm$1.26} & \textbf{72.74}\scalebox{0.8}{$\pm$1.04} & \textbf{71.59} & \textbf{80.77} \Tstrut \\
\bottomrule
\end{tabular}
}
\vspace{-3mm}
\end{table*}

\textbf{Average Accuracy after last task and Average Incremental Accuracy.}
Table~\ref{Tab:maintable} shows that our method MORE consistently outperforms the baselines. We compare with the replay-based methods first. The best replay-based method by average ACA and AIA over all the datasets is DER++. Our method achieves 71.59 and 80.77 in ACA and AIA, respectively, much better than 66.89 and 78.46 of DER++. This demonstrates that the existing replay-based methods utilizing the samples to prevent forgetting are inferior to our MORE using samples for OOD learning. The best baseline is the generative method PASS. Its average ACA and AIA over all the datasets are 68.25 and 77.11, respectively. Our method achieves much better performances of 71.59 and 80.77 on ACA and AIA, respectively. The performances of the multi-head method HAT~\citep{serra2018overcoming} using task-id prediction are 62.68 and 73.84 for ACA and AIA, respectively. These numbers are lower than many baselines. Its performance is particularly low in experiments where the number of classes per task is small. For instance, its ACA and AIA on CIFAR100-20T are 56.72 and 69.12, respectively, much lower than our method of 70.53 and 81.59 trained based on OOD detection. Thus, introducing the OOD term in the regularization in MORE improves it drastically.

\textbf{Classification Accuracy with Smaller Memory Sizes.}
For all the datasets, we run additional experiments with half of the original memory size and show that our method is even stronger with a smaller memory. The new memory sizes are 100, 1000, and 1000 for CIFAR-10, CIFAR-100, and Tiny-ImageNet, respectively. Table~\ref{Tab:smaller_memory} shows that
MORE has experienced almost no performance loss
with reduced memory size while the memory-based baselines suffer from performance reduction. The ACA and AIA values of the best memory-based baseline (DER++) have decreased from 66.89 to 62.16 and 78.46 to 75.40, respectively, while MORE only decreases from 71.59 to 71.44 and 80.77 to 79.61. The little reduction in performance is because a small number of OOD samples is enough to enable the system to produce a more robust network against samples beyond the current task.
\begin{table*}[t]
\caption{
Performance of the baselines and our method MORE with smaller memory sizes. We reduce the size of the memory buffer by half. The new sizes are 100, 1000, 1000 for CIFAR10, CIFAR100, and Tiny-ImageNet. Numbers in bold are the best results in each column.
}
\centering
\resizebox{0.96\columnwidth}{!}{
\begin{tabular}{l c c c c c c c c c c | c c}
&&&&&&&&&&&&\\[0.0em]
\toprule
\multirow{2}{*}{Method}  & \multicolumn{2}{c}{CIFAR10-5T}  &  \multicolumn{2}{c}{CIFAR100-10T} &  \multicolumn{2}{c}{CIFAR100-20T} &  \multicolumn{2}{c}{T-ImageNet-5T} & \multicolumn{2}{c}{T-ImageNet-10T} & \multicolumn{2}{|c}{Average}\\
{} & ACA & AIA & ACA & AIA & ACA & AIA & ACA & AIA & ACA & AIA & ACA & AIA \\
\midrule
OWM             &  41.69\scalebox{0.8}{$\pm$6.34}  & 59.07\scalebox{0.8}{$\pm$3.31} & 21.39\scalebox{0.8}{$\pm$3.18} & 39.71\scalebox{0.8}{$\pm$1.35} & 16.98\scalebox{0.8}{$\pm$4.44} & 32.18\scalebox{0.8}{$\pm$1.51} & 24.55\scalebox{0.8}{$\pm$2.48} & 45.65\scalebox{0.8}{$\pm$1.15} & 17.52\scalebox{0.8}{$\pm$3.45} & 35.57\scalebox{0.8}{$\pm$1.83} & 24.43 & 41.99 \Tstrut \\
iCaRL         & 86.08\scalebox{0.8}{$\pm$1.19}  & 92.18\scalebox{0.8}{$\pm$1.24}  &  66.96\scalebox{0.8}{$\pm$2.08}  & 76.70\scalebox{0.8}{$\pm$0.98}  & 68.16\scalebox{0.8}{$\pm$0.71}  & 77.78\scalebox{0.8}{$\pm$1.29}  & 47.27\scalebox{0.8}{$\pm$3.22}  & 59.95\scalebox{0.8}{$\pm$3.20}  & 49.51\scalebox{0.8}{$\pm$1.87}  & 62.02\scalebox{0.8}{$\pm$3.33}  & 63.60 & 73.72 \\ 
A-GEM         &  56.64\scalebox{0.8}{$\pm$4.29}  & 68.33\scalebox{0.8}{$\pm$1.73}  &  23.18\scalebox{0.8}{$\pm$2.54}  & 44.29\scalebox{0.8}{$\pm$1.01}  & 20.76\scalebox{0.8}{$\pm$2.88}  & 36.63\scalebox{0.8}{$\pm$0.61}  & 31.44\scalebox{0.8}{$\pm$3.84}  & 50.83\scalebox{0.8}{$\pm$2.76}  & 23.73\scalebox{0.8}{$\pm$6.27}  & 40.53\scalebox{0.8}{$\pm$3.72}  & 31.15 & 48.12 \\
EEIL & 77.44\scalebox{0.8}{$\pm$3.04} & 86.53\scalebox{0.8}{$\pm$1.20} & 62.95\scalebox{0.8}{$\pm$0.68} & 78.66\scalebox{0.8}{$\pm$0.53} & 57.86\scalebox{0.8}{$\pm$0.74} & 76.39\scalebox{0.8}{$\pm$0.59} & 48.36\scalebox{0.8}{$\pm$1.38} & 63.53\scalebox{0.8}{$\pm$0.85} & 44.59\scalebox{0.8}{$\pm$1.72} & 62.98\scalebox{0.8}{$\pm$0.39} & 58.24 & 73.62 \\
GD & 85.96\scalebox{0.8}{$\pm$1.64} & 92.75\scalebox{0.8}{$\pm$0.58} & 57.17\scalebox{0.8}{$\pm$1.06} & 76.57\scalebox{0.8}{$\pm$0.28} & 50.30\scalebox{0.8}{$\pm$0.58} & 73.36\scalebox{0.8}{$\pm$0.73} & 46.09\scalebox{0.8}{$\pm$1.77} & 64.18\scalebox{0.8}{$\pm$0.71} & 32.41\scalebox{0.8}{$\pm$2.75} & 57.68\scalebox{0.8}{$\pm$0.32} & 54.39 & 72.91 \\
DER++         &  80.09\scalebox{0.8}{$\pm$3.00}  & 89.42\scalebox{0.8}{$\pm$0.95} & 64.89\scalebox{0.8}{$\pm$2.48}  & 78.90\scalebox{0.8}{$\pm$0.26} & 65.84\scalebox{0.8}{$\pm$1.46}  & 79.58\scalebox{0.8}{$\pm$0.62} & 50.74\scalebox{0.8}{$\pm$2.41} & 65.19\scalebox{0.8}{$\pm$0.76} & 49.24\scalebox{0.8}{$\pm$5.01}  & 63.90\scalebox{0.8}{$\pm$0.67} & 62.16 & 75.40 \\
HAL           &  79.16\scalebox{0.8}{$\pm$4.56}  & 88.31\scalebox{0.8}{$\pm$1.06}  & 62.65\scalebox{0.8}{$\pm$0.83}  & 76.33\scalebox{0.8}{$\pm$0.67}  & 63.96\scalebox{0.8}{$\pm$1.49}  & 76.75\scalebox{0.8}{$\pm$0.84}  & 48.17\scalebox{0.8}{$\pm$2.94}  & 63.95\scalebox{0.8}{$\pm$1.64}  & 47.11\scalebox{0.8}{$\pm$6.00}  & 62.43\scalebox{0.8}{$\pm$1.74}  & 60.21 & 73.55 \\
PASS          &  86.21\scalebox{0.8}{$\pm$1.10}  & 91.78\scalebox{0.8}{$\pm$1.12}   & 68.90\scalebox{0.8}{$\pm$0.94}  & 78.27\scalebox{0.8}{$\pm$0.81}   & 66.77\scalebox{0.8}{$\pm$1.18}  & 77.01\scalebox{0.8}{$\pm$1.13}  & 61.03\scalebox{0.8}{$\pm$0.38} & 70.02\scalebox{0.8}{$\pm$0.56} & 58.34\scalebox{0.8}{$\pm$0.42}  & 68.45\scalebox{0.8}{$\pm$1.20} & 68.25 & 77.11 \\
HAT          &  83.30\scalebox{0.8}{$\pm$1.54}  & 91.06\scalebox{0.8}{$\pm$0.36}   & 62.34\scalebox{0.8}{$\pm$0.93}  & 73.99\scalebox{0.8}{$\pm$0.86}   & 56.72\scalebox{0.8}{$\pm$0.44}  & 69.12\scalebox{0.8}{$\pm$1.06}  & 57.91\scalebox{0.8}{$\pm$0.72} & 69.38\scalebox{0.8}{$\pm$1.14} & 53.12\scalebox{0.8}{$\pm$0.94}  & 65.63\scalebox{0.8}{$\pm$1.64} & 62.68 & 73.84 \\
\hline
MORE  &  \textbf{88.13}\scalebox{0.8}{$\pm$1.16}  & \textbf{93.82}\scalebox{0.8}{$\pm$0.28}  & \textbf{71.69}\scalebox{0.8}{$\pm$0.11}  & \textbf{79.83}\scalebox{0.8}{$\pm$0.49} & \textbf{71.29}\scalebox{0.8}{$\pm$0.55}  & \textbf{80.44}\scalebox{0.8}{$\pm$1.42} & \textbf{64.17}\scalebox{0.8}{$\pm$0.77}  & \textbf{72.48}\scalebox{0.8}{$\pm$0.25} & \textbf{61.90}\scalebox{0.8}{$\pm$0.90}  & \textbf{71.46}\scalebox{0.8}{$\pm$0.56} & \textbf{71.44} & \textbf{79.61} \Tstrut \\
\bottomrule
\end{tabular}
}
\label{Tab:smaller_memory}
\vspace{-3mm}
\end{table*}

\textbf{Average Performance Reduction Rate (Backward Transfer)}

\begin{wrapfigure}[15]{r}{0.3\textwidth}
\vspace{-8mm}
    \centering
    \includegraphics[width=1.8in]{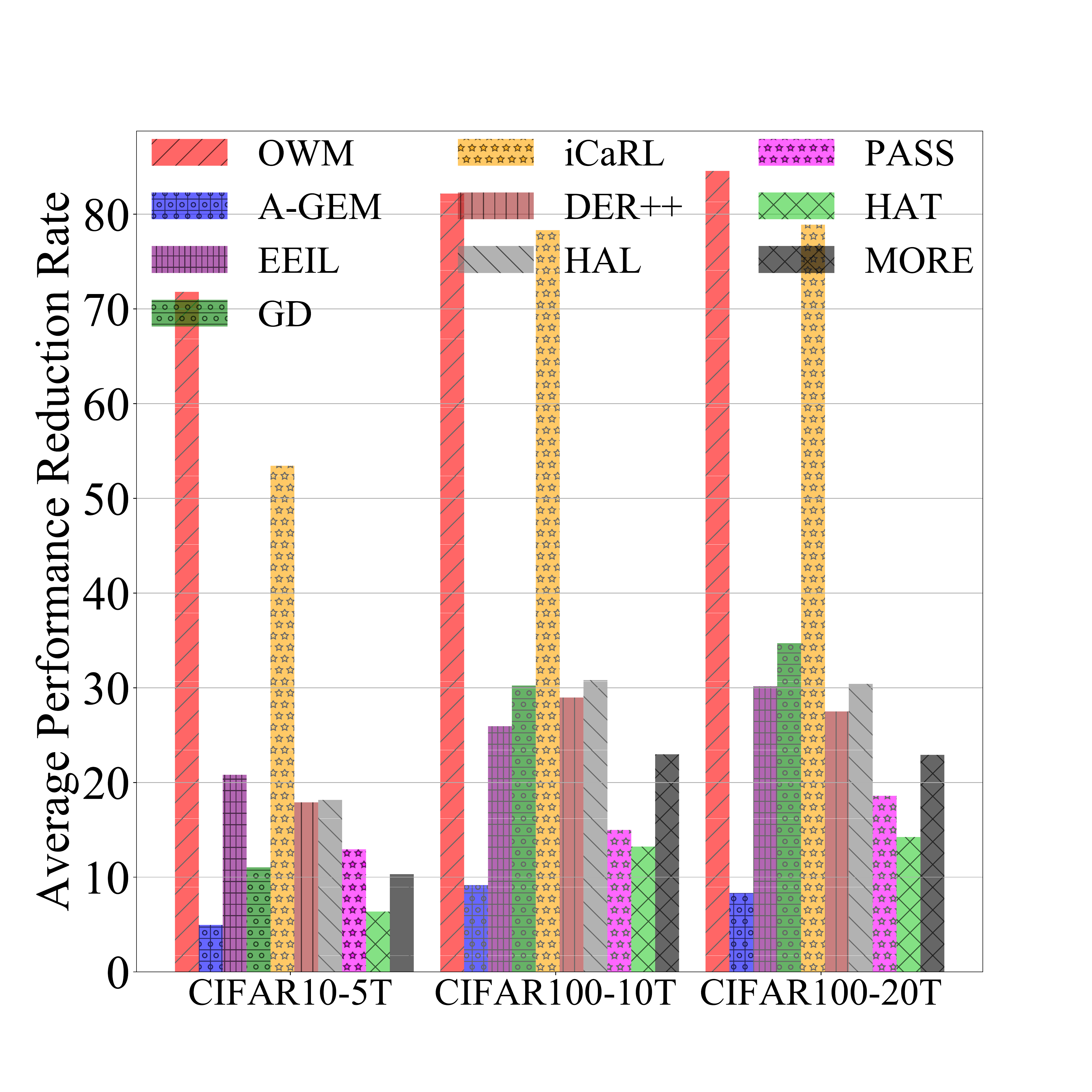}
    \caption{Average performance reduction rate (\%). The lower the rate, the better the method is.}
    \label{forgetting}
\end{wrapfigure}
We compare the performance reduction rate of our method against the baselines using CIFAR10-5T, CIFAR100-10T, and CIFAR100-20T. Figure~\ref{forgetting} shows that the performance of our method drops relatively lower than many baselines. A-GEM and HAT achieve lower reduction rate than our method in all the datasets. However, they are not able to adapt to new tasks well as ACA of A-GEM on the datasets after the final task are 56.33, 25.21, and 21.99, respectively, and those of HAT are 83.30, 62.34, and 56.72. The accuracy of our method on the same datasets are 89.16, 70.23, and 70.53. PASS experiences smaller drop in performance on CIFAR100-10T and CIFAR100-20T than our method, but its ACA of 68.90 and 66.77 are significantly lower than 70.23 and 70.53 of our MORE. For forward and backward transfer, refer to Appendix~\ref{appendix:forward_backward}.

As we explained in the introduction section, since our method MORE is based on OOD detection in building each task model, our method can naturally be used to detect test samples that are out-of-distribution for all the classes or tasks learned thus far. We are not aware of any existing continual learning system that has done such an evaluation. We evaluate the performance of the baselines and our system in this out-of distribution scenario, which is also called the open set setting.

\textbf{Out-of-Distribution Detection Results.}
This OOD detection ability is highly desirable for a continual learning system
because in a real-life environment, the system can be exposed to not only seen classes, but also unseen classes. When the test sample is from one of seen classes, the system should be able to predict its class. If the sample does not belong to any of the training classes seen so far (i.e., the sample is out-of-distribution), the system should detect it.

We formulate the performance of OOD detection of a continual learning system as the following. A continual learning system accepts and classifies a test sample after training task $k$ if the test sample is from one of the classes in tasks $1, \cdots, k$. If it is from one of the classes of the future tasks $k + 1, \cdots, t$, it should be rejected as OOD (where $t$ is the last task in each evaluation dataset). 

\begin{table*}[t]
\caption{AUC and incremental AUC (IAUC) after the second last task. The last task is not considered as there is no task after it and thus no OOD data after that. Numbers in bold are the best results in each column.}
\centering
\resizebox{0.96\columnwidth}{!}{
\begin{tabular}{l c c c c c c c c c c | c c}
&&&&&&&&&&&&\\[0.0em]
\toprule
\multirow{2}{*}{Method}  & \multicolumn{2}{c}{CIFAR10-5T}  &  \multicolumn{2}{c}{CIFAR100-10T} &  \multicolumn{2}{c}{CIFAR100-20T} &  \multicolumn{2}{c}{T-ImageNet-5T} & \multicolumn{2}{c}{T-ImageNet-10T} & \multicolumn{2}{|c}{Average}\\
{} & AUC & IAUC & AUC & IAUC & AUC & IAUC & AUC & IAUC & AUC & IAUC & AUC & IAUC \\
\midrule
OWM             &  58.26\scalebox{0.8}{$\pm$17.38}  & 70.02\scalebox{0.8}{$\pm$3.59}   & 50.87\scalebox{0.8}{$\pm$2.86}  & 63.17\scalebox{0.8}{$\pm$1.06}  & 55.43\scalebox{0.8}{$\pm$10.25}  & 59.42\scalebox{0.8}{$\pm$1.26} & 58.20\scalebox{0.8}{$\pm$2.51} & 67.24\scalebox{0.8}{$\pm$0.92} & 56.17\scalebox{0.8}{$\pm$4.26}  & 62.17\scalebox{0.8}{$\pm$0.35} & 55.79 & 64.41 \Tstrut \\
iCaRL         & 78.54\scalebox{0.8}{$\pm$9.59}  & 82.12\scalebox{0.8}{$\pm$5.38}  &  72.10\scalebox{0.8}{$\pm$2.66}  & 77.42\scalebox{0.8}{$\pm$0.45}  & 69.79\scalebox{0.8}{$\pm$5.75}  & 76.91\scalebox{0.8}{$\pm$1.30} & 66.05\scalebox{0.8}{$\pm$1.73} & 71.86\scalebox{0.8}{$\pm$1.57}  & 66.62\scalebox{0.8}{$\pm$1.77}  & 74.24\scalebox{0.8}{$\pm$1.66} & 70.19 & 76.06 \\ 
A-GEM         &  63.71\scalebox{0.8}{$\pm$15.18}  & 74.92\scalebox{0.8}{$\pm$5.62}   &  52.18\scalebox{0.8}{$\pm$2.60} & 64.19\scalebox{0.8}{$\pm$0.86}  & 54.78\scalebox{0.8}{$\pm$13.40}  & 60.23\scalebox{0.8}{$\pm$0.95} & 58.97\scalebox{0.8}{$\pm$2.52} & 67.88\scalebox{0.8}{$\pm$1.28}  & 56.33\scalebox{0.8}{$\pm$4.14}  & 63.08\scalebox{0.8}{$\pm$1.12} & 57.19 & 66.06 \\
EEIL & 81.56\scalebox{0.8}{$\pm$10.62} & 87.19\scalebox{0.8}{$\pm$2.31} & 67.39\scalebox{0.8}{$\pm$3.44} & 78.89\scalebox{0.8}{$\pm$1.32} & 64.83\scalebox{0.8}{$\pm$8.01} & 77.69\scalebox{0.8}{$\pm$1.40} & 67.22\scalebox{0.8}{$\pm$2.16} & 74.82\scalebox{0.8}{$\pm$0.79} & 62.36\scalebox{0.8}{$\pm$6.14} & 73.45\scalebox{0.8}{$\pm$1.33} & 68.58 & 78.39 \\
GD & \textbf{85.02}\scalebox{0.8}{$\pm$9.88} & \textbf{89.71}\scalebox{0.8}{$\pm$1.85} & 64.22\scalebox{0.8}{$\pm$2.47} & 77.31\scalebox{0.8}{$\pm$1.03} & 61.95\scalebox{0.8}{$\pm$9.02} & 75.19\scalebox{0.8}{$\pm$0.87} & 68.35\scalebox{0.8}{$\pm$2.97} & 75.36\scalebox{0.8}{$\pm$0.78} & 58.79\scalebox{0.8}{$\pm$3.10} & 70.90\scalebox{0.8}{$\pm$1.75} & 67.67 & 77.69 \\
DER++         &  79.25\scalebox{0.8}{$\pm$4.74}  & 84.61\scalebox{0.8}{$\pm$2.64}  & 70.36\scalebox{0.8}{$\pm$1.81}  & 78.42\scalebox{0.8}{$\pm$0.64}  & 69.74\scalebox{0.8}{$\pm$2.02}  & 78.37\scalebox{0.8}{$\pm$0.42}  & 68.67\scalebox{0.8}{$\pm$3.83} & 74.80\scalebox{0.8}{$\pm$1.72} & 67.81\scalebox{0.8}{$\pm$0.23}  & 74.86\scalebox{0.8}{$\pm$1.93} & 70.93 & 78.09 \\
HAL           &  77.97\scalebox{0.8}{$\pm$9.76}  & 84.09\scalebox{0.8}{$\pm$3.30}  & 69.55\scalebox{0.8}{$\pm$0.83}  & 77.37\scalebox{0.8}{$\pm$0.55}  & 71.58\scalebox{0.8}{$\pm$3.54}  & 77.66\scalebox{0.8}{$\pm$0.31}  & 67.58\scalebox{0.8}{$\pm$3.71}  & 74.52\scalebox{0.8}{$\pm$1.93}  & 67.27\scalebox{0.8}{$\pm$1.86}  & 75.47\scalebox{0.8}{$\pm$2.35}  & 70.79 & 77.82 \\
PASS          &  77.69\scalebox{0.8}{$\pm$4.01}  & 84.57\scalebox{0.8}{$\pm$1.54}   & 71.80\scalebox{0.8}{$\pm$2.41}  & 77.74\scalebox{0.8}{$\pm$1.40}   & 66.62\scalebox{0.8}{$\pm$5.78}  & 77.42\scalebox{0.8}{$\pm$1.44}  & 71.61\scalebox{0.8}{$\pm$1.10} & 77.07\scalebox{0.8}{$\pm$2.14} & 68.51\scalebox{0.8}{$\pm$4.49}  & 74.79\scalebox{0.8}{$\pm$2.36} & 71.24 & 78.32 \\
HAT          &  83.89\scalebox{0.8}{$\pm$4.10}  & 87.83\scalebox{0.8}{$\pm$2.44}   & 71.26\scalebox{0.8}{$\pm$1.93}  & 79.57\scalebox{0.8}{$\pm$0.29}   & 65.52\scalebox{0.8}{$\pm$3.43}  & 77.20\scalebox{0.8}{$\pm$0.74}  & 75.08\scalebox{0.8}{$\pm$1.07} & 79.78\scalebox{0.8}{$\pm$1.59} & 72.02\scalebox{0.8}{$\pm$1.35}  & 78.25\scalebox{0.8}{$\pm$1.68} & 73.55 & 80.53 \\
\hline
MORE & 80.83\scalebox{0.8}{$\pm$8.82} & 88.06\scalebox{0.8}{$\pm$1.84} & \textbf{73.32}\scalebox{0.8}{$\pm$2.80} & \textbf{81.67}\scalebox{0.8}{$\pm$1.27} & \textbf{72.28}\scalebox{0.8}{$\pm$4.81} & \textbf{80.97}\scalebox{0.8}{$\pm$0.80} & \textbf{75.74}\scalebox{0.8}{$\pm$2.66} & \textbf{80.72}\scalebox{0.8}{$\pm$3.38} & \textbf{72.78}\scalebox{0.8}{$\pm$1.08} & \textbf{79.73}\scalebox{0.8}{$\pm$2.97} & \textbf{74.99} & \textbf{82.23} \Tstrut \\
\bottomrule
\end{tabular}
}
\label{Tab:maintable_auc}
 \vspace{-3mm}
\end{table*}

We use maximum softmax probability (MSP)~\citep{hendrycks2016baseline_msp} as the OOD score of a test sample for the baselines and use maximum output with coefficient in Eq.~\ref{final_prediction} for our method. We employ
Area Under the ROC Curve (AUC) to measure the performance of OOD detection as AUC is the standard metric used in OOD detection papers~\citep{yang2021generalized}. We report the AUC after the second last task. We do not consider the AUC after the last task because there is no more continual learning task (and thus, no OOD) after the final task. We also report incremental AUC (IAUC) which is the average of AUC over all the tasks.

Table~\ref{Tab:maintable_auc} shows that our method consistently outperforms the baselines in all the datasets.
The best baselines based on ACA and AIA are DER++ and PASS. Their average AUC and IAUC over the experiments are 70.93 and 78.09 for DER++ and 71.24 and 78.32 for PASS. Our method MORE achieves 74.99 and 82.23. The performance gain in OOD detection problem is not because of better performance in classification. For example, on CIFAR100-10T and CIFAR100-20T, the AIA of DER++ is slightly better than MORE in Table~\ref{Tab:maintable} with memory size 2000. However, in IAUC, DER++ achieves 78.42 and 78.37 for CIFAR100-10T and 20T, respectively, while MORE obtains 81.67 and 80.97. The improvement is due to OOD detection in training.

\subsection{Ablation Study}
We conduct an ablation study to measure the performance gain by each proposed technique: back-update
in previous models in Sec.~\ref{sec:backward} and the distance-based coefficient in Sec.~\ref{sec:ensemble_scores} using three experiments. The back-update by Eq.~\ref{backward} is to improve the earlier task models as they are trained with less diverse OOD data than later models. The modified output with coefficient in Eq.~\ref{eq:ensemble} is to improve the classification accuracy by assembling or combining the softmax probability of the task networks and the inverse Mahalanobis distances.

Table~\ref{ablation:tab} compares
AIA
obtained after applying each method. Both distance based coefficient and back-update show large improvements from the original method without any of the two techniques. Although the performance is already competitive with either technique, the performance improves further after applying them together.

\begin{wraptable}[10]{r}{2.8in}
\vspace{-0.15in}
\caption{
Performance gains with the proposed techniques. The row Original indicates the method without the coefficient and back-update and the row Back means the back-update method.
}
\centering
\resizebox{2.8in}{!}{
\begin{tabular}{l c c c}
\toprule
{} & \multicolumn{1}{c}{CIFAR10-5T} & \multicolumn{1}{c}{CIFAR100-10T} & \multicolumn{1}{c}{CIFAR100-20T}\\
\midrule
\multicolumn{1}{c}{Original} & 91.01\scalebox{0.8}{$\pm$2.48} & 76.93\scalebox{0.8}{$\pm$1.58} & 75.76\scalebox{0.8}{$\pm$2.35}  \\
\multicolumn{1}{c}{Coefficient (C)} & 93.86\scalebox{0.8}{$\pm$1.12} & 80.31\scalebox{0.8}{$\pm$1.02} & 80.77\scalebox{0.8}{$\pm$1.36}  \\
\multicolumn{1}{c}{Back (B)} & 93.36\scalebox{0.8}{$\pm$0.79} & 80.35\scalebox{0.8}{$\pm$1.08} & 80.32\scalebox{0.8}{$\pm$0.82} \\
\midrule
\multicolumn{1}{c}{C + B} & 94.23\scalebox{0.8}{$\pm$0.82} & 81.24\scalebox{0.8}{$\pm$1.24} & 81.59\scalebox{0.8}{$\pm$0.98} \\
\bottomrule
\end{tabular}
}
\label{ablation:tab}
\end{wraptable}
As discussed in Sec.~\ref{sec:backward}, the earlier tasks are trained with less diverse OOD samples than later tasks because the samples in memory are considered as OOD data in training. As a result, the models suffer from over-confidence problem in later tasks, which is a phenomenon that a neural network produces a high score for OOD instance. For example, in our experiment with CIFAR100-20T, we found that after training the last task, 73\% of incorrectly classified test instances are classified as one of the classes in earlier tasks than the true tasks while only 27\% of incorrectly classified samples are predicted as one of the classes in later tasks. The average accuracy is 66.19. After updating the previous networks using the back-update method, 
equal proportion (49.72\%) of incorrectly classified samples are predicted to either earlier or later tasks
and the accuracy increases to 70.53. The back-update helps eliminate the asymmetric performance issue in the task models and improves the classification accuracy.

\subsection{Additional Experiments}

\textbf{Number of Pre-Training Classes.}
We study the effect of the number of classes of the data used in pre-training the feature extractor. For the above experiments, we used 611 classes of ImageNet after removing 389 classes similar to the classes in CIFAR10/100 and Tiny-ImageNet to pre-train the feature extractor. Here, we randomly select 200 classes from the 611 classes and use the data to pre-train a feature extractor. The results are reported in Table~\ref{Tab:maintable_pre_train_200}. The average ACA and AIA (last column) over 5 experiments show that our method MORE still outperforms the baselines. Despite that some baselines outperform MORE in some cases, the differences are small. For the more challenging dataset, Tiny-ImageNet, MORE is markedly superior to the baselines.

\textbf{Performance of Pre-Trained Feature Extractor.}
We provide the performance of the pre-trained feature extractor to measure how much value the proposed method add to the feature extractor.
We fix the pre-trained feature extractor and construct a classifier on top of the feature representations of the training data. The considered classifiers are nearest class mean
classifier (NCM) and cosine similarity classifier (CSC). We use the mean (i.e., prototype) of the class features to represent each class in the classifier. Given a test instance, we use Euclidean distance (or cosine similarity) for NCM (for CSC) between the test sample and the prototypes, and choose the class that gives the minimum distance (or maximum similarity). Since these classifiers incrementally add new prototypes without training the classifier or feature extractor, they are not continual learning methods, but are appropriate to evaluate the performance of a pre-trained feature extractor.
We report the results of CIFAR10, CIFAR100, and
Tiny-ImageNet by using all the samples in a single task.
NCM achieves 85.80, 58.94, and 54.75 on the three datasets, respectively. CSC achieves 85.93, 58.97, 54.75. On the other hand, our method MORE achieves 89.16, 70.23/70.53, and 64.97/63.06 on CIFAR10-5T, CIFAR100-10T/20T, and T-ImageNet-5T/10T, respectively. Our MORE is considerably better, which shows that despite the strong feature extractor, one can achieve much better result by training a model with our MORE method than simply building a classifier on top of the pre-trained feature extractor.

\begin{table*}[t]
\caption{
Average classification accuracy (ACA) and average incremental accuracy (AIA) after the final task. `-XT' means X number of tasks. The results are based on a pre-trained model using 200 randomly selected classes from the 611 classes of ImageNet after removing 389 classes similar to the classes in CIFAR10/100 and Tiny-ImageNet. Our system MORE and all baselines used the pre-trained network. The last two columns show the average ACA and AIA of each method over all datasets and experiments. We highlight the best results in each column in bold.
}
\centering
\resizebox{0.96\columnwidth}{!}{
\begin{tabular}{l c c c c c c c c c c | c c}
&&&&&&&&&&&&\\[0.0em]
\toprule
\multirow{2}{*}{Method}  & \multicolumn{2}{c}{CIFAR10-5T}  &  \multicolumn{2}{c}{CIFAR100-10T} &  \multicolumn{2}{c}{CIFAR100-20T} &  \multicolumn{2}{c}{T-ImageNet-5T} & \multicolumn{2}{c}{T-ImageNet-10T} & \multicolumn{2}{|c}{Average}\\
{} & ACA & AIA & ACA & AIA & ACA & AIA & ACA & AIA & ACA & AIA & ACA & AIA \\
\midrule
OWM             & 30.04\scalebox{0.8}{$\pm$9.98} & 49.88\scalebox{0.8}{$\pm$3.67} & 11.51\scalebox{0.8}{$\pm$5.63} & 30.78\scalebox{0.8}{$\pm$1.75} & 17.68\scalebox{0.8}{$\pm$4.66} & 24.28\scalebox{0.8}{$\pm$0.95} & 8.57\scalebox{0.8}{$\pm$3.85} & 33.92\scalebox{0.8}{$\pm$1.70} & 10.24\scalebox{0.8}{$\pm$5.18} & 26.47\scalebox{0.8}{$\pm$4.47}& 15.61 & 33.07 \Tstrut \\
iCaRL          & 77.51\scalebox{0.8}{$\pm$0.77} & 85.87\scalebox{0.8}{$\pm$1.13} & 59.83\scalebox{0.8}{$\pm$0.64} & 69.79\scalebox{0.8}{$\pm$1.24} & \textbf{58.27}\scalebox{0.8}{$\pm$0.23} & 70.00\scalebox{0.8}{$\pm$1.14} & 42.85\scalebox{0.8}{$\pm$0.77} & 53.30\scalebox{0.8}{$\pm$0.90} & 42.11\scalebox{0.8}{$\pm$0.57} & 54.61\scalebox{0.8}{$\pm$1.29}& 56.11 & 66.71 \\
A-GEM          & 47.70\scalebox{0.8}{$\pm$0.98} & 62.04\scalebox{0.8}{$\pm$1.87} & 16.15\scalebox{0.8}{$\pm$2.00} & 34.67\scalebox{0.8}{$\pm$0.57} & 9.78\scalebox{0.8}{$\pm$0.64} & 24.20\scalebox{0.8}{$\pm$0.39} & 21.26\scalebox{0.8}{$\pm$1.03} & 39.71\scalebox{0.8}{$\pm$0.67} & 13.56\scalebox{0.8}{$\pm$2.44} & 30.13\scalebox{0.8}{$\pm$1.06}& 21.69 & 38.15 \\
EEIL & 67.50\scalebox{0.8}{$\pm$4.46} & 81.42\scalebox{0.8}{$\pm$0.60} & \textbf{59.41}\scalebox{0.8}{$\pm$0.36} & \textbf{74.48}\scalebox{0.8}{$\pm$0.72} & 55.17\scalebox{0.8}{$\pm$0.83} & \textbf{72.83}\scalebox{0.8}{$\pm$0.91} & 43.10\scalebox{0.8}{$\pm$0.47} & 57.12\scalebox{0.8}{$\pm$0.41} & 40.63\scalebox{0.8}{$\pm$0.59} & 56.76\scalebox{0.8}{$\pm$0.53} & 53.16 & 68.52 \\
GD  & \textbf{80.44}\scalebox{0.8}{$\pm$0.86} & \textbf{88.99}\scalebox{0.8}{$\pm$0.85} & 55.88\scalebox{0.8}{$\pm$0.71} & 73.89\scalebox{0.8}{$\pm$0.60} & 50.42\scalebox{0.8}{$\pm$0.52} & 70.41\scalebox{0.8}{$\pm$0.76} & 43.01\scalebox{0.8}{$\pm$0.90} & 58.45\scalebox{0.8}{$\pm$0.56} & 32.47\scalebox{0.8}{$\pm$1.78} & 54.28\scalebox{0.8}{$\pm$0.54} & 52.44 & 69.21 \\
DER++         & 69.11\scalebox{0.8}{$\pm$1.24} & 81.60\scalebox{0.8}{$\pm$1.33} & 56.02\scalebox{0.8}{$\pm$0.73} & 71.77\scalebox{0.8}{$\pm$0.53} & 56.22\scalebox{0.8}{$\pm$0.41} & 72.08\scalebox{0.8}{$\pm$0.83} & 40.67\scalebox{0.8}{$\pm$1.17} & 54.36\scalebox{0.8}{$\pm$0.37} & 38.51\scalebox{0.8}{$\pm$2.15} & 53.78\scalebox{0.8}{$\pm$0.21}& 52.11 & 66.72 \\
HAL           & 68.13\scalebox{0.8}{$\pm$2.63} & 80.10\scalebox{0.8}{$\pm$1.33} & 53.56\scalebox{0.8}{$\pm$0.75} & 67.54\scalebox{0.8}{$\pm$0.55} & 53.56\scalebox{0.8}{$\pm$1.11} & 66.86\scalebox{0.8}{$\pm$0.82} & 38.52\scalebox{0.8}{$\pm$2.13} & 54.54\scalebox{0.8}{$\pm$0.75} & 35.98\scalebox{0.8}{$\pm$3.57} & 50.75\scalebox{0.8}{$\pm$0.85}& 49.95 & 63.96 \\
PASS          & 76.92\scalebox{0.8}{$\pm$0.60} & 85.98\scalebox{0.8}{$\pm$1.30} & 57.67\scalebox{0.8}{$\pm$0.64} & 69.72\scalebox{0.8}{$\pm$0.77} & 54.64\scalebox{0.8}{$\pm$0.83} & 67.41\scalebox{0.8}{$\pm$1.01} & 48.51\scalebox{0.8}{$\pm$0.54} & 58.92\scalebox{0.8}{$\pm$0.85} & 45.01\scalebox{0.8}{$\pm$0.53} & 56.77\scalebox{0.8}{$\pm$1.17}& 56.55 & 67.76 \\
HAT          & 67.50\scalebox{0.8}{$\pm$3.75} & 80.33\scalebox{0.8}{$\pm$1.06} & 51.08\scalebox{0.8}{$\pm$0.23} & 64.22\scalebox{0.8}{$\pm$0.72} & 46.84\scalebox{0.8}{$\pm$0.57} & 61.54\scalebox{0.8}{$\pm$1.16} & 50.54\scalebox{0.8}{$\pm$0.77} & 60.80\scalebox{0.8}{$\pm$1.16} & 45.68\scalebox{0.8}{$\pm$0.54} & 57.91\scalebox{0.8}{$\pm$1.34}& 52.33 & 64.96 \\
\hline
MORE            & 72.56\scalebox{0.8}{$\pm$6.41} & 85.23\scalebox{0.8}{$\pm$3.23} & 58.68\scalebox{0.8}{$\pm$1.90} & 70.78\scalebox{0.8}{$\pm$0.65} & 58.02\scalebox{0.8}{$\pm$0.66} & 69.79\scalebox{0.8}{$\pm$1.34} & \textbf{51.75}\scalebox{0.8}{$\pm$0.32} & \textbf{61.86}\scalebox{0.8}{$\pm$0.66} & \textbf{48.00}\scalebox{0.8}{$\pm$0.89} & \textbf{58.96}\scalebox{0.8}{$\pm$0.68}& \textbf{57.80} & \textbf{69.32} \Tstrut \\
\bottomrule
\end{tabular}
}
\label{Tab:maintable_pre_train_200}
 \vspace{-3mm}
\end{table*}

\section{Conclusion}
This paper proposes a novel approach MORE for class incremental learning (CIL). Unlike the existing memory-based CIL methods that use a single-head model and leveraging the memory buffer to mitigate forgetting, the proposed method uses a multi-head model and a memory buffer for building a task specific OOD model for each task which can identify the class of a sample from the task and also detect OOD samples. Since the parameters are effectively protected by task identifiers in the multi-head setting through hard attentions, the network successfully adapts to new knowledge with little reduction in performance. The resulting network outperforms the state-of-the-art baselines in terms of standard continual learning metrics and also shows strong performance in OOD detection in the continual learning setting.

\newpage
\section*{Acknowledgments}
{\color{black}This work was supported in part by a research contract from KDDI, two National Science Foundation (NSF) grants (IIS-1910424 and IIS-1838770), a DARPA contract HR001120C0023, and a Northrop Grumman research gift.}

\bibliography{collas2022_conference}
\bibliographystyle{collas2022_conference}

\appendix
\section{Pseudo-Code} \label{appendix:pseudo}
For task $k$, Let $p(y | \vx, k) = \text{softmax} f(h(\vx, k; \theta, \ve^{k}); \phi_k)$, where $\theta$ is the parameters for adapter, $\ve^{k}$ is the trainable embedding for hard attentions, and $\phi_{k}$ is the set of parameters of the classification head of task $k$. Algorithm~\ref{MOREtrain} and Algorithm~\ref{MOREpred} describe the training and testing processes, respectively. We add comments with the symbol ``//''.
\begin{algorithm}[H]
\caption{Training MORE}\label{MOREtrain}
\begin{algorithmic}[1]
    \Require Memory $\mathcal{M}$, learning rate $\lambda$, a sequence of tasks $\mathcal{D}=\{\mathcal{D}^k \}_{k=1}$, and parameters $\{\theta, \ve, \phi \}$,
    where $\ve$ and $\phi$ are collections of task embeddings $\ve^{k}$ and task heads $\phi_{k}$
    \Statex // CL starts
    \For{each task data $\mathcal{D}^k \in \mathcal{D}$}
        \Statex \hspace{0.5cm} // Model training
        \For{a batch $(\mX_{i}^{k}, \vy)$ in $\mathcal{D}^k$, until converge}
            \State $\mX_s$ = $sample(\mathcal{M})$
            \State Compute $\mathcal{L}$ of Eq.~\ref{final_obj} and gradients of parameters
            \State Modify the model parameters $\nabla \theta \leftarrow \nabla \theta'$ using Eq.~\ref{eq:grad_mod}
            \State Update parameters as $\theta \leftarrow \theta - \lambda \nabla \theta, \ \ve^{k} \leftarrow \ve^{k} - \lambda \partial \mathcal{L}, \ \phi_{k} \leftarrow \phi_{k} - \lambda \partial \mathcal{L}$ 
        \EndFor
        \Statex \hspace{0.5cm} // Back-updating in Sec.~\ref{sec:backward}
        \State Randomly select $\tilde{\mathcal{D}} \subset \mathcal{D}^{k}$, where $|\tilde{\mathcal{D}}| = |\mathcal{M}|$
        \For{each task $j$, until converge}
            \State minimize $\mathcal{L}(\phi_{j})$ of Eq.~\ref{backward}
        \EndFor
        \Statex \hspace{0.5cm} // Obtain statistics in Sec.~\ref{sec:ensemble_scores}
        \State Compute $\vmu^k_{j}$ using Eq.~\ref{eq:compute_mu} and $\mS^{k}$ using Eq.~\ref{eq:compute_sigma}
    \EndFor

\end{algorithmic}
\end{algorithm}

\begin{algorithm}[H]
\caption{MORE Prediction}\label{MOREpred}
\begin{algorithmic}[1]
    \Require Test instance $\vx$ and parameters $\{\theta, \ve, \phi \}$
    \For {each task $k$}
    \State Obtain $p(\mathcal{Y}^{k} | \vx, k)$
    \State Obtain $s^{k}(\vx)$ using Eq.~\ref{eq:ensemble}
    \EndFor
    \Statex // Concatenate outputs for final prediction $y$ and OOD score $s$
    \State $y = \argmax \bigoplus_{1\leq k \leq t}  p(\mathcal{Y}^{k} | \vx, k) s^{k}(\vx)$ (i.e. Eq.~\ref{final_prediction})
    \State $s = \max \bigoplus_{1\leq k \leq t}  p(\mathcal{Y}^{k} | \vx, k) s^{k}(\vx)$
\end{algorithmic}
\end{algorithm}

\section{Large Distribution Shift from Pre-Training to CL datasets}\label{appendix:additional_data}
This section evaluates model performance when the distribution between the dataset used for pre-training and the dataset used in continual learning is very different. \cite{kumar2022finetuning} have showed that fine-tuning does not work well for OOD detection if this data distribution shift is large. We use street view house numbers (SVHN)~\citep{netzer2011reading_svhn} to show the effect of a large distribution shift from the pre-training dataset. Our feature extractor is pre-trained with the ImageNet dataset, which does not contain any class of data similar to house numbers. 

We split the SVHN dataset into 5 tasks, where each task consists of 2 classes. We train DER++ and PASS, the two best performing baselines in our experiment. The average classification accuracy values after the final task are 68.19, 76.09 for DER++ and PASS, respectively. Our method achieves 80.57. Despite the fact that the distribution of the pre-training dataset and the continual learning dataset are very different, our method still outperforms them.

\section{Size of Memory Required}\label{appendix:size_of_memory}
In this section, we report sizes of memory required by each method. The sizes include network size, replay buffer, all other parameters or example kept in memory simultaneously for a model to be functional.

We use an `entry' to refer to a parameter or element in a vector or matrix to calculate the total memory required to train and test. The pre-trained backbone uses 21.6 million (M) entries (parameters). The adapter modules use 1.2M entries for CIFAR10 and 2.4M for other datasets. The baselines and our method use 22.8M and 24.0M entries for the model on CIFAR10 and other datasets, respectively. The unique technique of each method may add additional entries for training and test/inference. 

The total memory required for each method without considering the replay memory buffer is reported in Table~\ref{Tab:memory}. Our method is competitive in memory consumption. Baselines such as OWM and A-GEM take considerably more memory than our system. iCaRL and DER++ take the least amount of memory, but the differences between our method and theirs are only 0.8M, 1.8M, 3.6M, 1.0M, and 1.8M for CIFAR10, CIFAR100-10T, CIFAR100-20T, T-ImageNet-5T and T-ImageNet-10T.

\begin{table*}[t]
\caption{
Total memory (in entries) required for each method without the replay memory buffer.
}
\centering
\resizebox{0.8\columnwidth}{!}{
\begin{tabular}{l c c c c c}
\toprule
\multirow{1}{*}{Method}  & \multicolumn{1}{c}{CIFAR10-5T}  &  \multicolumn{1}{c}{CIFAR100-10T} &  \multicolumn{1}{c}{CIFAR100-20T} &  \multicolumn{1}{c}{T-ImageNet-5T} & \multicolumn{1}{c}{T-ImageNet-10T} \\
\midrule
OWM & 26.6M & 28.1M & 28.1M & 28.2M & 28.2M \Tstrut \\
iCaRL & 22.9M & 24.1M & 24.1M & 24.1M & 24.1M \\
A-GEM & 26.5M & 31.4M & 31.4M & 31.5M & 31.5M \\
EEIL & 22.9M & 24.1M & 24.1M & 24.1M & 24.1M \\
GD & 22.9M & 24.1M & 24.1M & 24.1M & 24.1M \\
DER++ & 22.9M & 24.1M & 24.1M & 24.1M & 24.1M \\
HAL & 22.9M & 24.1M & 24.1M & 24.1M & 24.1M \\
PASS & 22.9M & 24.2M & 24.2M & 24.3M & 24.4M \\
HAT & 23.0M & 24.7M & 25.4M & 24.6M & 25.1M \\
MORE & 23.7M & 25.9M & 27.7M & 25.1M & 25.9M  \\
\bottomrule
\end{tabular}
}
\label{Tab:memory}
 \vspace{-3mm}
\end{table*}

Many replay based methods (e.g., iCaRL, HAL) need to save the previous network for distillation during training. This requires additional 1.2M or 2.4M entries for CIFAR10 or other datasets. Our method does not save the previous model as we do not use distillation.

Note that a large memory consumption usually comes from the memory buffer as the raw data is of size 32*32*3 or 64*64*3 for CIFAR and T-ImageNet. For memory buffer of size 2000, a system needs 6.1M or 24.6M entries for CIFAR or T-ImageNet. Therefore, saving a smaller number of samples is important for reducing the memory consumption. As we demonstrated in Table~\ref{Tab:maintable} and Table~\ref{Tab:smaller_memory} in the main paper, our method performs better than the baselines even with a smaller memory buffer. In Table~\ref{Tab:maintable}, we use large memory sizes (e.g., 200 and 2000 for CIFAR10 and other datasets). In Table~\ref{Tab:smaller_memory}, we reduce the memory size by half. When we compare the accuracy of our method in Table~\ref{Tab:smaller_memory} to those of the baselines in Table~\ref{Tab:maintable}, our method still outperforms them on all datasets. Our method with a smaller memory buffer achieves average classification accuracy of 88.13, 71.69, 71.29, 64.17, 61.90 on CIFAR10-5T, CIFAR100-10T, CIFAR100-20T, T-ImageNet-5T, and T-ImageNet-10T. On the other hand, the best baselines achieve 88.98, 69.73, 70.03, 61.03, 58.34 on the same experiments with a larger memory buffer.

\section{Forward and Backward Transfer}\label{appendix:forward_backward}
We follow \citep{NEURIPS2020_cat} for forward transfer (FWT) and \citep{Lopez2017gradient} for backward transfer (BWT). FWT is basically the average performance difference on each task between a CL method when the task is first learned and a trained individual model from random initialization. By definition, the higher the FWT, the better the transfer is. For our experiments, we train the pre-trained model with adapters instead of a randomly initialized model. The FWTs of the three best baselines (iCaRL, DER++, PASS) and our method on CIFAR100-10T are -19.04, -1.4, -13.79, and -5.28. The FWTs on Tiny-ImageNet are -17.33, 2.96, -11.18, and -10.12. DER++ shows the best FWT, but its final accuracy values are lower than MORE due to bad backward transfer or forgetting (see below).

The BWT in \citep{Lopez2017gradient} is the same as the performance reduction rate that we report in the paper. We report the exact values here. The BWT of the three best baselines (iCaRL, DER++, PASS) and our method are -8.32, -26.51, -18.37, and -19.75 on CIFAR100-10T. The BWT on Tiny-ImageNet-5T are -10.69, -25.70, -7.14, -10.96. 
iCaRL and PASS show better BWT (lower reduction in performance from the initial accuracy) than MORE. However, MORE achieves better average classification accuracies than the baselines as discussed in Sec.~\ref{sec:result_comparison}. These baselines are good at preserving the past knowledge, but their learning methods are weak at learning each task, or vice versa.
Our method shows good balance between FWT and BWT and achieve the best average accuracy.
\end{document}